\documentclass{article}

\usepackage[textheight=9in, 
	textwidth=5.5in,
	top=1in,
	headheight=12pt,
	headsep=25pt,
    footskip=30pt]{geometry}
\usepackage{sectsty}
\sectionfont{\fontsize{12}{14}\selectfont}
\subsectionfont{\fontsize{11}{13}\selectfont}

\usepackage[utf8]{inputenc} 
\usepackage[T1]{fontenc}    
\usepackage{hyperref}       
\usepackage{url}            
\usepackage[numbers, compress]{natbib}
\usepackage{booktabs}       
\usepackage{amsfonts}       
\usepackage{nicefrac}       
\usepackage{microtype}      
\usepackage{graphicx}  
\usepackage{booktabs}
\usepackage{xcolor}
\usepackage{amssymb,amsmath}
\usepackage{subcaption}

\newcommand{\ba}{\boldsymbol{a}}
\newcommand{\rba}{\widetilde{\ba}}

\DeclareMathOperator*{\smin}{smin}
\DeclareMathOperator*{\sargmin}{sargmin}

\title{The Clever Hans Effect in Anomaly Detection}

\author{%
  \normalsize Jacob Kauffmann\textsuperscript{\textnormal 1}, Lukas Ruff\textsuperscript{\textnormal 1}, Gr\'egoire Montavon\textsuperscript{\textnormal 1}, Klaus-Robert M{\"u}ller\textsuperscript{\textnormal 1,2,3,4} \\
  \ \\
  \normalsize \textsuperscript{1} Technische Universit{\"a}t Berlin, Machine Learning Group\\
  \normalsize \textsuperscript{2} Google Research, Brain Team, Berlin\\
  \normalsize \textsuperscript{3} Max Planck Institut f{\"u}r Informatik\\
  \normalsize \textsuperscript{4} Korea University\\
}
\date{}

\begin{document}

\maketitle

\begin{abstract}
The `Clever Hans' effect occurs when the learned model produces correct predictions based on the `wrong' features. This effect which undermines the generalization capability of an ML model and goes undetected by standard validation techniques has been frequently observed for supervised learning where the training algorithm leverages spurious correlations in the data. The question whether Clever Hans also occurs in unsupervised learning, and in which form, has received so far almost no attention. Therefore, this paper will contribute an explainable AI (XAI) procedure that can highlight the relevant features used by popular anomaly detection models of different type. Our analysis reveals that the Clever Hans effect is widespread in anomaly detection and occurs in many (unexpected) forms. Interestingly, the observed Clever Hans effects are in this case not so much due to the data, but due to the anomaly detection models themselves whose structure makes them unable to detect the truly relevant features, even though vast amounts of data points are available. Overall, our work contributes a warning against an unrestrained use of existing anomaly detection models in practical applications, but it also points at a possible way out of the Clever Hans dilemma, specifically, by allowing multiple anomaly models to mutually cancel their individual structural weaknesses to jointly produce a better and more trustworthy anomaly detector.
\end{abstract}

\section{Introduction}

Anomaly detection \cite{chandola2009anomaly,pimentel2014} is a common machine learning problem for which numerous approaches have been proposed. They can be roughly organized into density-based \cite{Parzen1962,pidhorskyi2018}, reconstruction-based \cite{hawkins2002,hoffmann2007,DBLP:conf/cvpr/CongYL11},
and boundary-based approaches \cite{DBLP:journals/neco/ScholkopfPSSW01,DBLP:journals/ml/TaxD04,schlegl2017,DBLP:conf/icml/RuffGDSVBMK18}. Recently, there has been an effort to develop benchmarks and procedures to systematically assess and compare the performance of different anomaly detection models \cite{Goldstein2016, DBLP:conf/cvpr/BergmannFSS19}. Because it is hard to conceive a representative set of anomalies that would test for all particular ways in which a point can be anomalous, common testing procedures (based on a score and some set of labeled outliers) can strongly under/overestimate the performance of a given trained anomaly detection model.

In this paper, we propose to further validate an anomaly detector by inspecting its decision structure, in particular, to ensure that the model has not unintentionally implemented a `Clever Hans' strategy \cite{Lapuschkin2019}. `Clever Hans' refers to a famous horse that was believed to be able to perform arithmetic calculations asked by his trainer. Later studies have revealed, that the horse was basing its consistently correct predictions not on performing the actual mathematical  calculation but on watching unintended gestures of the trainer or some other human. Similar Clever Hans effects have been observed in areas such as information retrieval \cite{DBLP:journals/tmm/Sturm14} or supervised learning \cite{Lapuschkin2019}. In the latter case, this typically arises from letting the learning model exploit spurious correlations in data. These spurious correlations have been predicted to become increasingly severe as datasets become larger \cite{Calude2016}, thus, making the problem increasingly relevant. To the best of our knowledge, however, this interesting effect has not yet been studied in unsupervised models such as anomaly detection.

To analyze the Clever Hans effect in the context of anomaly detection, we need to endow the anomaly detection models with explainable AI (XAI) capabilities, so that they are able to provide explanations to their own prediction, specifically, identifying input features that have contributed the most to the anomaly score. Various methods have been proposed for explaining anomaly models \cite{DBLP:conf/icdm/MicenkovaNDA13,DBLP:conf/gpce/KowalAT16,DBLP:journals/pr/KauffmannMM20}, although so far, explanation techniques do not homogeneously and reliably apply to all anomaly detection models. Also, there is a need to systematically detect Clever Hans effects from those explanations, something that has so far only been done for supervised classifiers \cite{Lapuschkin2019}.

In this paper, we address the questions above by contributing in the following distinct ways:
\begin{enumerate}
\item We propose to embed common anomaly detection models in a three-layer neural network architecture consisting of (i) feature extraction, (ii) distance computation and (iii) pooling. This common representation of anomaly models allows us to systematically explain their predictions in a way that is comparable across models.
\item We demonstrate in the context of anomaly detection that `Clever Hans' effects can also be found in large amounts and in various forms. For this, we make use of the MNIST-C \cite{DBLP:journals/corr/abs-1906-02337} and MVTec \cite{DBLP:conf/cvpr/BergmannFSS19} corpora which come with ground-truth anomalies and pixel-wise annotations.
\item We analyze on a more abstract level how the Clever Hans effect systematically arises in different anomaly detection models. Unlike in supervised learning, where this effect is mainly the result of exploiting spurious correlations in the data, Clever Hans is found here to be intrinsic to the {\em structure} of the anomaly detection model.
\item Observing that each model has structural weaknesses that lead to Clever Hans strategies, we investigate a bagging approach, which we find to produce improvements compared to the individual models.
\end{enumerate}

Overall, through the lens of XAI, our paper brings attention to the limitations of standard anomaly detection models, in particular, their exposure to the Clever Hans effect. While it underlines further motivation and necessity for improving current anomaly detectors, it also shows that the Clever Hans effect can be reduced via a simple bagging technique.

\section{Towards Explainable Anomaly Detection}
\label{section:models}

In the following, we introduce a common XAI framework that is applicable to a broad range of anomaly detection models by first embedding them into a layered neural network architecture, and then using the resulting layered architecture to support a common backpropagation procedure for extracting explanations. Anomaly detection models can be roughly divided into three categories:

\paragraph{Density-Based Models} Density-based anomaly detection consists of first learning a probability model on inlier data, and then measure outlierness for a new data point $x$ based on the model probability score, e.g.\ $o(x) = -\log p_\theta(x)$. For the common {\em kernel density estimation} models \cite{Rosenblatt1956,Parzen1962} (KDE), the outlier score can be rewritten up to a constant factor and additive term as a soft min-pooling over distances
\begin{align}
o(x) = \textstyle \min_j^\gamma ( \|x - x_j\|^2).
\label{eq:kde}
\end{align}
When the kernel is Gaussian, the min-pooling is given as $\min_j^\gamma (h_j) = -\gamma^{-1}\log\sum_j\exp(-\gamma \, h_j)$, i.e.\ an inverted log-sum-exp pooling with stiffness $\gamma$. When the kernel is t-Student, the min-pooling takes the form of a harmonic mean \cite{DBLP:journals/pr/KauffmannMM20}. A similar min-pooling formulation can be obtained for other types of density models such as Gaussian mixture models.

\paragraph{Reconstruction-Based Models} Reconstruction is another paradigm for anomaly detection \cite{hawkins2002, hoffmann2007, DBLP:conf/cvpr/CongYL11}, where the outlier score is given as the divergence between the original data and its reconstruction, for example given by an {\em autoencoder} $x \mapsto r(x)$. Typically, the outlier score is given by the squared reconstruction distance:
$$
o(x) = \|r(x)-x\|^2.
$$
When the autoencoder output is not deterministic and produces instead a conditional distribution $p_\theta(x\,|\,r(x))$ at the output, outlierness can be measured as the negative log-probability, and when the conditional distribution is Gaussian, it becomes equivalent to the squared distance.

\paragraph{Boundary-Based Models} Boundary-based models learn an envelope that contains the inliers. Data points outside the envelope are then predicted to be outliers. The envelope or decision boundary can be built in input space, e.g.\ as in the one-class SVM \cite{DBLP:journals/neco/ScholkopfPSSW01} or Support Vector Data Description \cite{DBLP:journals/ml/TaxD04}, but also in feature space, e.g.\ in deep one-class classification \cite{DBLP:conf/icml/RuffGDSVBMK18}. A deep one-class model typically builds a spherical envelope in feature space by defining the outlier score
\begin{align*}
o(x) = \|\phi(x)\|^2
\end{align*}
where the feature map is trained to minimize this quantity for inliers, subject to some regularization constraint. Interestingly, boundary-based models are not restricted to unsupervised data and can instead leverage supervised outlier data to refine the decision boundary \cite{DBLP:journals/jair/GoernitzKRB13,DBLP:conf/accv/AkcayAB18,ruff2020}.

\medskip

\begin{figure}[h]
\centering \includegraphics[width=\textwidth]{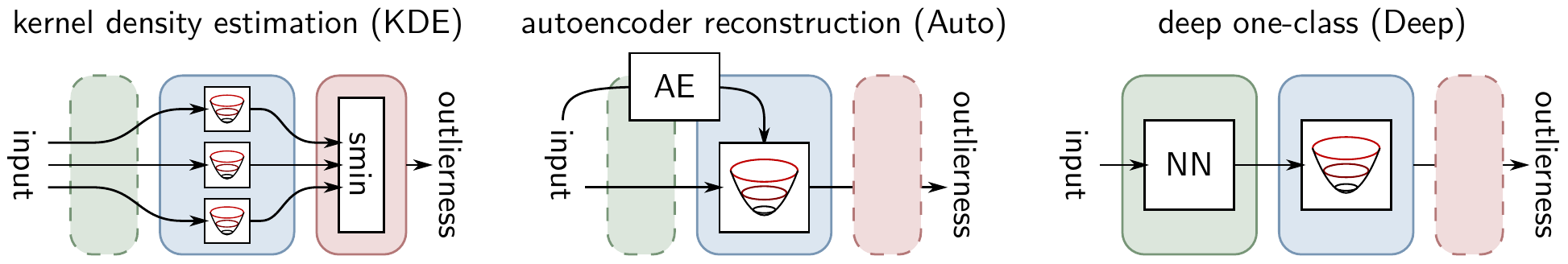}
\caption{Three-layer view (features\,/\,distance\,/\,pooling) of anomaly detection models of interest.}
\label{figure:layers}
\end{figure}

We have briefly reviewed three families of methods for outlier detection: density-based, reconstruction-based, and boundary-based. Although these methods may appear to be different in all technical aspects, we find that outlier scores produced by each of them can in fact be conceptually embedded in the same three-layer architecture: (1) {\em Feature extraction:} Features that are considered to be relevant for the task of anomaly detection are extracted or made more salient in the feature map representation. (This step is only present in the deep one-class model). (2) {\em Distances:} Distances to some template(s) are then produced. Templates are either the data distribution itself for the KDE model, the autoencoder reconstruction for the reconstruction-based model, or the origin in feature space for the deep one-class model. (3) {\em Pooling:} Produced distances are then pooled using some common pooling function (this step is only present in the KDE model). The way each model fits in this three-layer architecture is shown in Figure \ref{figure:layers}.

This architecture lets us build an explanation technique that homogeneously applies to all three models. Our approach to explanation is based on the framework of `Deep Taylor Decomposition' \cite{DBLP:journals/pr/MontavonLBSM17,DBLP:journals/pr/KauffmannMM20}, where a Taylor expansion is applied at each layer to identify the most contributing terms in the lower layer. Overall, the method leads to an LRP-type \cite{bach-plos15} backward propagation procedure from the output score to the input features. Propagation in the different layers is detailed below.

\paragraph{Pooling Layer} Let $o((s_k)_k)$  be some function performing the pooling. We identify a `deactivation' line in the space of pooled elements on which the pooling function is linear. We then select the root point $(\widetilde{s}_k)_k$ on that line, i.e.\ where the output of the pooling becomes zero. Finally, the attribution is given by the elements of a first-order Taylor expansion at the root point, i.e.\ $R_k = [\nabla o(\widetilde{s})]_k \cdot (s_k - \widetilde{s}_k).$ To further propagate to the lower layers, we can approximate the attribution as a homogeneous function of the pooled activations, i.e.\ $\widehat{R}_k = c_k s_k$.

\paragraph{Distance Layer} When attribution hits the distance layer, the output must be further redistributed to different dimensions entering in the distance computation. Here, we express the score we have obtained in the layer above in terms of the distance inputs: $\widehat{R}_k(a) = c_k \sum_j (a_{jk} - \mu_{jk})^2$. The variable $\mu_{jk}$ is either constant (KDE and Deep) or is treated as such (Autoencoder). A second-order Taylor expansion at the origin gives diagonal second-order terms: $R_{jk} = c_k \cdot (a_{jk} -\mu_{jk})^2$. 

\paragraph{Features Layer} To further propagate the decomposition throughout the network, we need $\mu=0$ (i.e. the outlier model is centered at the origin), which lets us express the terms from the layer above as $R_{jk} = c_k \phi_{jk}(x)^2$. This quantity can be propagated to the input features using standard LRP \cite{bach-plos15}.

Further explanations and justification for these propagation steps are given in Appendix \ref{appx:explain_details} of the Supplement. Finally, when compared to other approaches, especially sampling-based approaches, which do not rely on the underlying model structure \cite{DBLP:conf/icdm/MicenkovaNDA13,DBLP:conf/kdd/Ribeiro0G16}, the propagation-based approach we have developed here is faster (it requires only a single pass in the model) and also does not require access to the input distribution.

\section{Unmasking Clever Hans Effects in Anomaly Models}
\label{section:benchmark}

Having equipped anomaly detection models with a systematic way of explaining their predictions in terms of input features, we will now look for possible Clever Hans effects in these anomaly models, especially when the latter are applied to real data.

Our experiments are performed on two recent datasets: MNIST-C \cite{DBLP:journals/corr/abs-1906-02337}, and MVTec \cite{DBLP:conf/cvpr/BergmannFSS19}. A peculiarity of these two datasets which makes them ideal testbeds is that they either come with the data generation process (from which ground-truth explanations of anomalies can be built) or with ground-truth pixel-wise annotations of the anomaly patterns. 

For each dataset, we train three models: A kernel density estimator (KDE), an autoencoder (Auto), and a deep one-class model (Deep). For the KDE model, we use a Gaussian kernel where we select the scale such that the likelihood of an inlier validation set is maximized. For the autoencoder on MNIST-C, we use a LeNet-type encoder that has two convolutional layers with max-pooling followed by two fully connected layers that map to an encoding of 64 dimensions. We construct the decoder symmetrically where we replace convolution and max-pooling with deconvolution and upsampling respectively. For MVTec, we use an encoder-decoder architecture as presented in \cite{huang2019inverse} which maps to a bottleneck of 512 dimensions. Both, the encoder and decoder here consist of four blocks having two $3{\times}3$ convolutional layers followed by max-pooling or upsampling respectively. We train the autoencoders using the Adam optimizer \cite{kingma2014} such that the reconstruction error of an inlier validation set is minimized. 
The Deep One-Class model is built on top of the feature extractor of a CNN. For MNIST-C, we train a classifier CNN on the `letters' subset of the EMNIST dataset \cite{DBLP:journals/corr/CohenATS17}. The feature extractor consists of three convolutional layers that are interleaved with $\ell_2$ pooling layers, leading to a 640 dimensional feature space. For MVTec, we start with a standard pretrained VGG-16 network \cite{DBLP:journals/corr/SimonyanZ14a} and we cut off the top classification layer, which results in a 4096 dimensional feature space. For both models, the outlier score is defined to be the squared norm of the feature vector after a linear whitening transformation that is regularized such that the outlier detection ROC score on a validation set with some outliers is maximized.

The detection performance for each class, model and dataset are shown in Figure \ref{figure:detection}. 

\begin{figure}[h]
\small
\begin{minipage}{0.5\textwidth}
\centering
\textbf{MNIST-C}\\[2mm]
\begin{tabular}{rrrr}\toprule
& KDE & Auto & Deep\\\midrule
          brightness & \bf 100.0 & \bf 100.0 &      13.7\\
         canny edges &      78.9 & \bf 100.0 &      97.9\\
         dotted line &      68.5 & \bf  99.9 &      86.4\\
                 fog &      62.1 & \bf 100.0 &      17.4\\
          glass blur &       8.0 & \bf  99.4 &      31.1\\
       impulse noise &      98.0 & \bf 100.0 &      97.5\\
         motion blur &       8.1 & \bf  94.1 &      70.7\\
              rotate &      37.1 &      53.8 & \bf  65.5\\
               scale &       5.0 &      39.4 & \bf  79.8\\
               shear &      49.9 & \bf  69.6 &      64.6\\
          shot noise &      41.6 & \bf  99.5 &      51.5\\
             spatter &      44.5 & \bf  96.8 &      68.2\\
              stripe & \bf 100.0 & \bf 100.0 &     100.0\\
           translate &      76.2 &      90.2 & \bf  98.8\\
              zigzag &      84.0 & \bf 100.0 &      94.3\\
\bottomrule\end{tabular}

\end{minipage}%
\begin{minipage}{0.5\textwidth}
\centering
\textbf{MVTec}\\[2mm]
\begin{tabular}{rrrr}\toprule
& KDE & Auto & Deep\\\midrule
              bottle &      83.3 &      95.0 & \bf  99.6\\
               cable &      66.9 &      57.3 & \bf  90.9\\
             capsule &      56.2 &      52.5 & \bf  91.0\\
              carpet &      34.8 &      36.8 & \bf  90.6\\
                grid &      71.7 & \bf  74.6 &      52.4\\
            hazelnut &      69.9 &      90.5 & \bf  95.0\\
             leather &      41.5 &      64.0 & \bf  78.3\\
           metal nut &      33.3 &      45.5 & \bf  85.2\\
                pill &      69.1 &      76.0 & \bf  80.4\\
               screw &      36.9 &      77.9 & \bf  86.9\\
                tile &      68.9 &      51.8 & \bf  96.5\\
          toothbrush &      93.3 &      49.4 & \bf  96.4\\
          transistor &      72.4 &      51.2 & \bf  90.8\\
                wood & \bf  94.7 &      88.5 &      91.6\\
              zipper &      61.4 &      35.0 & \bf  92.4\\
\bottomrule\end{tabular}

\end{minipage}
\caption{ROC detection performance for every model on each class of MNIST-C and MVTec.}
\label{figure:detection}
\end{figure}

On the MNIST-C dataset, the autoencoder delivers the best results, however, there are exceptions to this, for example, scale, rotation, or translations are better detected by the deep one-class model. On the MVTec dataset, the deep one-class model performs best on most classes, which can be explained by the more abstract level and multiscale properties present in this dataset. The question we ask is whether the results are truly reliable or whether some of the accuracies are compromised due to the Clever Hans effect.

Here, we leverage the fact that for the two considered datasets, MNIST-C and MVTec, ground-truth pixel-wise annotations are given and can therefore be confronted with the pixel-wise explanations of the model predictions. We can define a score measuring the mismatch between the {\em detection accuracy} (measured as the area under the ROC curve) and the {\em explanation accuracy} (measured as the cosine similarity between the ground-truth and the pixel-wise explanation). Pixel-wise explanations are passed through a rectification function so that the cosine similarity is always greater or equal to zero. We define the Clever Hans score as the difference:
$$
\text{Clever Hans score}~~=~~\text{detection accuracy}~~-~~\text{explanation accuracy}
$$
The Clever Hans score is a number which ranges from $-1$ to $1$ (or $-100$ and $100$ if expressed on a percentage scale). The closer to $100$, the more the anomaly detector has fooled the standard validation procedure by exploiting the wrong input features. In particular, we will look for anomaly detection models and tasks with highest Clever Hans scores to highlight the widespread presence of this effect in anomaly detection. Figure \ref{figure:top_classes} shows the top-3 classes with highest Clever Hans scores for each model and dataset.

\begin{figure}[h]
\small
\centering

\begin{minipage}{.3\textwidth}\centering KDE, MNIST-C \\\begin{tabular}{p{0.6\textwidth}p{0.2\textwidth}}\toprule
1. dotted line          & 31.9 \\
2. zigzag               & 31.4 \\
3. spatter              & 31.0 \\
\bottomrule\end{tabular}\end{minipage}
 ~~~~~~~~~~
\begin{minipage}{.3\textwidth}\centering KDE, MVTec \\\begin{tabular}{p{0.6\textwidth}p{0.2\textwidth}}\toprule
1. wood                 & 62.6 \\
2. grid                 & 61.4 \\
3. zipper               & 53.7 \\
\bottomrule\end{tabular}\end{minipage}

\bigskip

\begin{minipage}{.3\textwidth}\centering Auto, MNIST-C \\\begin{tabular}{p{0.6\textwidth}p{0.2\textwidth}}\toprule
1. shear                & 43.2 \\
2. canny edges          & 41.4 \\
3. motion blur          & 38.6 \\
\bottomrule\end{tabular}\end{minipage}
 ~~~~~~~~~~
\begin{minipage}{.3\textwidth}\centering Auto, MVTec \\\begin{tabular}{p{0.6\textwidth}p{0.2\textwidth}}\toprule
1. bottle               & 69.1 \\
2. grid                 & 66.4 \\
3. wood                 & 64.8 \\
\bottomrule\end{tabular}\end{minipage}

\bigskip

\begin{minipage}{.3\textwidth}\centering Deep, MNIST-C \\\begin{tabular}{p{0.6\textwidth}p{0.2\textwidth}}\toprule
1. stripe               & 59.7 \\
2. dotted line          & 48.4 \\
3. impulse noise        & 48.0 \\
\bottomrule\end{tabular}\end{minipage}
 ~~~~~~~~~~
\begin{minipage}{.3\textwidth}\centering Deep, MVTec \\\begin{tabular}{p{0.6\textwidth}p{0.2\textwidth}}\toprule
1. toothbrush           & 76.0 \\
2. screw                & 75.3 \\
3. zipper               & 74.9 \\
\bottomrule\end{tabular}\end{minipage}

\caption{Top-3 classes with highest Clever Hans scores for each model and dataset.}
\label{figure:top_classes}
\end{figure}

\begin{figure}[th]
\centering

\parbox{.11\textwidth}{\flushright \vskip -1.5cm \small KDE, MNIST-C (dotted line)}~~~\fbox{\includegraphics[width=.11\textwidth]{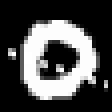}}
\includegraphics[width=.11\textwidth]{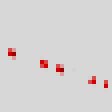}
\includegraphics[width=.11\textwidth]{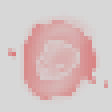}
\hfill
\parbox{.11\textwidth}{\flushright \vskip -1.5cm \small KDE, MVTec (wood)}~~~\fbox{\includegraphics[width=.11\textwidth]{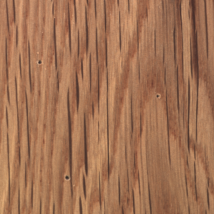}}
\includegraphics[width=.11\textwidth]{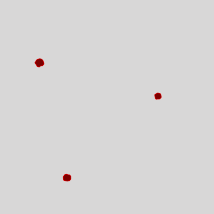}
\includegraphics[width=.11\textwidth]{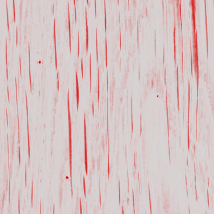}

\bigskip

\parbox{.11\textwidth}{\flushright \vskip -1.5cm \small Auto, MNIST-C (canny edges)}~~~\fbox{\includegraphics[width=.11\textwidth]{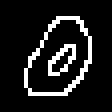}}
\includegraphics[width=.11\textwidth]{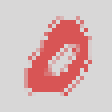}
\includegraphics[width=.11\textwidth]{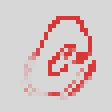}
\hfill
\parbox{.11\textwidth}{\flushright \vskip -1.5cm \small Auto, MVTec (bottle)}~~~\fbox{\includegraphics[width=.11\textwidth]{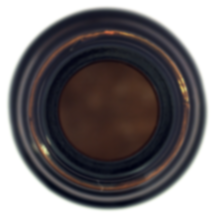}}
\includegraphics[width=.11\textwidth]{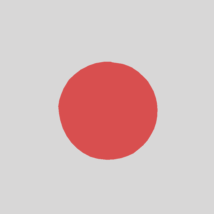}
\includegraphics[width=.11\textwidth]{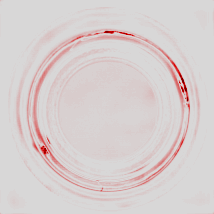}

\bigskip

\parbox{.11\textwidth}{\flushright \vskip -1.5cm \small Deep, MNIST-C (stripes)}~~~\fbox{\includegraphics[width=.11\textwidth]{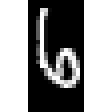}}
\includegraphics[width=.11\textwidth]{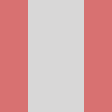}
\includegraphics[width=.11\textwidth]{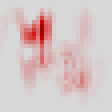}
\hfill
\parbox{.11\textwidth}{\flushright \vskip -1.5cm \small Deep, MVTec (zipper)}~~~\fbox{\includegraphics[width=.11\textwidth]{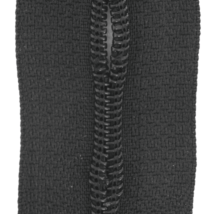}}
\includegraphics[width=.11\textwidth]{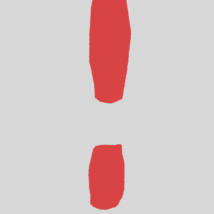}
\includegraphics[width=.11\textwidth]{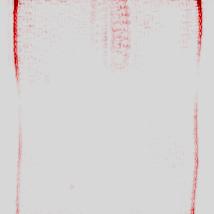}

\medskip

\caption{Examples taken for each model and dataset from one of the top-3 classes with highest Clever Hans score. For each case, we show from left to right, the input image, the ground-truth explanation, and the model-based explanation.}
\label{figure:heatmaps}
\end{figure}

Highest Clever Hans scores are obtained for the MVTec dataset which is also a more difficult and high-dimensional anomaly detection problem. It is also notable that different classes appear in the top-3 for the different models, suggesting that the models are affected by the problem in different ways. To shed light on the diverse Clever Hans effects, we look at single instances from classes in the top-3. Examples for each model and dataset are given in Figure \ref{figure:heatmaps}.

\setlength{\fboxsep}{0em}

We observe for example on MNIST-C, that the KDE model, although correctly identifying the anomalous dotted pattern, also highlights the whole digit region. The same occurs for the wood class on the MVTec dataset, where the high-frequency wood stripes appear as anomalous and completely dominate the small local perforations on the wood that are the true sources of anomaly.

Similar Clever Hans effects can be observed for the autoencoder, in particular, for the MNIST-C canny edges transformation. Here, although the whole interior of the digit has turned from white to black, the autoencoder completely ignores that change of color and only highlights the contour of the digit. On MVTec, a large contamination is present at the center of the bottle (photographed from above), however, the autoencoder bases its anomaly prediction on very fine elements on the outer part of the bottle.

Clever Hans effects can also be observed for the deep one-class model. The MNIST-C stripe transformation makes the whole border of the image turn from black to white, however, the deep one-class model bases its decision only on the edges of the added stripes and the interaction between these edges and the digit. On the MVTec data, the decision is mostly based on looking at the transition between the zipper  tissue and the white background, rather than attending the true source of anomaly which is the zipper opening.

In all the examples above, the anomaly model has produced high outlier scores, but these high scores were produced systematically for the `wrong' reasons. These Clever Hans strategies potentially undermine the generalization capability of the models, even for the classes with the highest measured anomaly detection accuracy.

\section{Understanding Why Anomaly Models are Clever Hanses}
\label{section:understanding}

In practice, it is not feasible to inspect the explanation of every new prediction. We thus need to attain more systematic insights into what conceptually and theoretically causes these Clever Hans strategies. Because the observed effects manifest themselves in various ways for the different models, we hypothesize that they are inherent to the structure of the anomaly detection models, rather than a petty effect of the training data. In the following, we give explanations for why the Clever Hans effect arises systematically in these models, and connect them to the effects observed in Fig.\ \ref{figure:heatmaps}.

\paragraph{Kernel Density Estimation:} We propose an explanation of the Clever Hans effect in KDE, based on the {\em concentration of distances} phenomenon occurring in high-dimensional spaces \cite{DBLP:journals/sadm/ZimekSK12}. In such spaces, distances between different pairs of data points (assumed to be sampled randomly) become increasingly similar, i.e.\ their deviation from the mean converges to zero. By rewriting the KDE model of Eq.\ \eqref{eq:kde} as
$$
\textstyle o(x) = \bar{d} -\gamma^{-1} \log \sum_{j=1}^N \exp (- \gamma (\|x-x_j\|^2 - \bar{d}))
$$
where $\bar{d}$ indicates the average distance to the inliers, and making use of the linear approximations $\exp(t) = 1+t$ and $\log(N+t) = \log(N) + t/N$ at $t=0$, the KDE outlier score can be approximated as
$$
o(x) \approx \|x-\bar{x}\|^2 - \log(N)/\gamma + \text{cst.}
$$
where $\bar{x}$ denotes the mean of the inliers (cf. Appendix \ref{appx:high-dim_kde} in the Supplement for a derivation). This result suggests that in effect, KDE implements a difference-to-the-mean anomaly detection strategy in the input space. This strategy is also revealed by the heatmaps in Fig.\ \ref{figure:heatmaps} where the difference-to-the-mean component (the digit area and the wood stripes) explains a greater fraction of the KDE outlier score than to the true anomalies (the dotted line and the wood perforations).

\paragraph{Autoencoder Reconstruction:} For this model, the Clever Hans effect finds its source in data points $x$ whose reconstruction $r(x)$ lies far away from the true data distribution. Hence, the difference $x-r(x)$ which supports the construction of the outlier score, is mainly supported by features that do not connect in any meaningful way to the input distribution and are therefore irrelevant for explaining anomaly. This weakness is also revealed by the heatmap explanations in Fig.\ \ref{figure:heatmaps}. The MNIST-C canny-edge transformation brings the data point in a completely different region of the input space, where the reconstruction-based explanation focuses on edges rather than focusing on the global transformation of the digit from white to black. A similar effect is observed on the MVTec data for the class bottle, where the large contamination artefact at the center of the image can be more easily reconstructed than some non-anomalous patterns at the border of the bottle, therefore, misidentifying again the true source of outlierness.

\paragraph{Deep One-Class:} Deep models measure outlierness as the distance in some feature space different from the input space $o(x) = \|\phi(x)-\phi(x')\|^2$. Typically, the feature map incorporates feature weightings, activation units, and pooling steps that produce distortions on the local input geometry. While such distortions are desirable in supervised learning to build invariance and gain statistical efficiency, it almost surely also compresses a few of the many components in which an anomaly could occur. This phenomenon can be clearly seen in the explanations of Fig.\ \ref{figure:heatmaps}: For the MNIST-C example, the outer border of the explanation is zero although the stripe outlier pattern extends to the very border of the image. Similarly, on the MVTec example, the deep network attends the border of the zipper fabric much more strongly than the center area where the true anomaly can be found. 

\begin{figure}[h]
\centering
\includegraphics[width=0.8\textwidth]{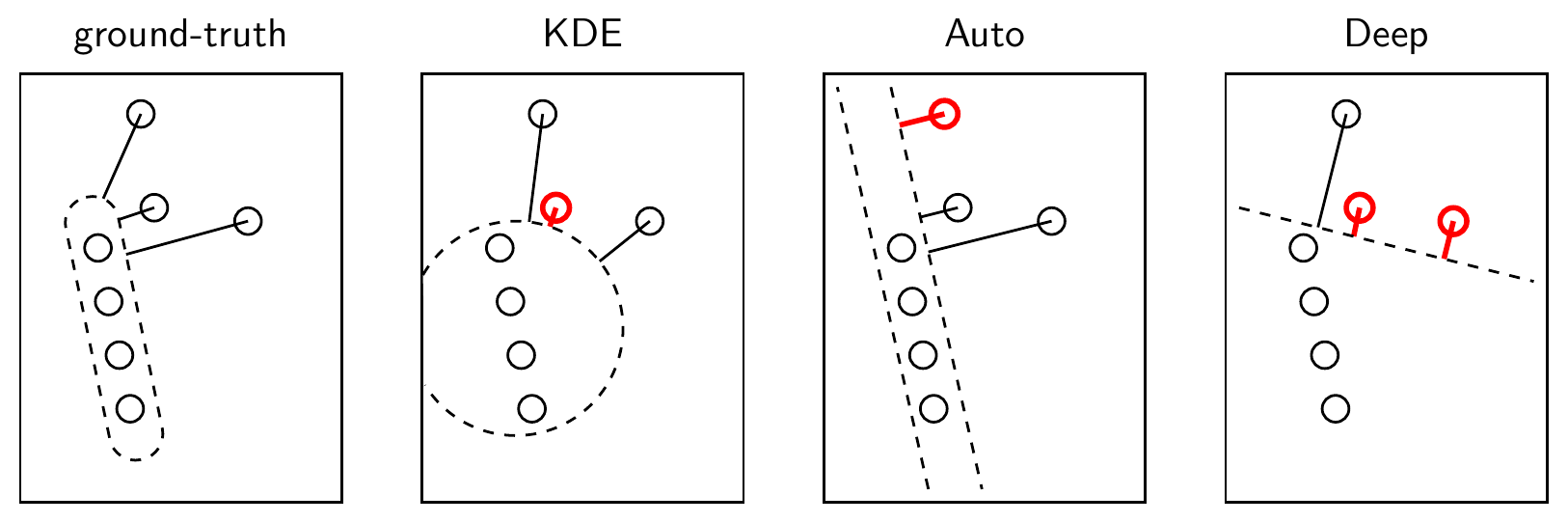}
\caption{Cartoon depiction of the weaknesses of different anomaly models. All outliers are predicted correctly. Clever Hans examples are shown in red.}
\label{figure:cartoon}
\end{figure}

The weaknesses of these different anomaly detection techniques and the Clever Hans strategies they induce are summarized as a cartoon in Fig.\ \ref{figure:cartoon}. The KDE model builds a hypersphere centered at the data mean. The autoencoder builds some reconstruction manifold (here from top to bottom) and the outlier score is given as the distance to that manifold. Finally, the deep model learns a separating surface between the inliers and outliers, whose orientation is strongly affected by the learned feature map.

While in this cartoon example all the data is correctly predicted, the explanations, here depicted as an orthogonal projection on the decision boundary, are strongly influenced by the structure of the model. Clever Hans predictions are highlighted in red and correspond to examples whose explanation deviates significantly from the ground-truth explanation.

To summarize, we have argued that flaws on the decision structure revealed by our explanation technique are mostly the consequence of model limitations and biases. This finding substantially differs from the study of Clever Hans in the context of supervised learning where it was found that such effect was rather induced by spurious correlations in the data \cite{Lapuschkin2019}.



\section{Improving Prediction with a Bag of Clever Hanses}

Motivated by the structural weaknesses of individual anomaly detection models, we discuss a simple improvement consisting of bagging these individual detectors to arrive at a better and more robust prediction strategy. Several works on bagging anomaly detectors have delivered encouraging results \cite{DBLP:conf/kdd/LazarevicK05,DBLP:conf/dasfaa/VuAG10}. The bagged model we consider here computes
$$
o_\text{Bag}(x) = \frac13 \big( o_\text{KDE}(x) + o_\text{Auto}(x) + o_\text{Deep}(x)\big),
$$
where the outlier scores of each individual model have been standardized over the training examples of the given class to have mean $0$ and variance $1$. The bagging can be understood as a soft max-pooling over outlier scores. It effectively implements a disjunction (i.e.\ logical `OR') of all outlier areas of the individual models. A geometrical motivation for this strategy can be found in Figure \ref{figure:cartoon}, where the resulting area would closely match the ground-truth outlier area shown on the left.

\begin{figure}[h]
\centering \small
\begin{minipage}{0.35\textwidth}
\centering
\includegraphics[width=\textwidth]{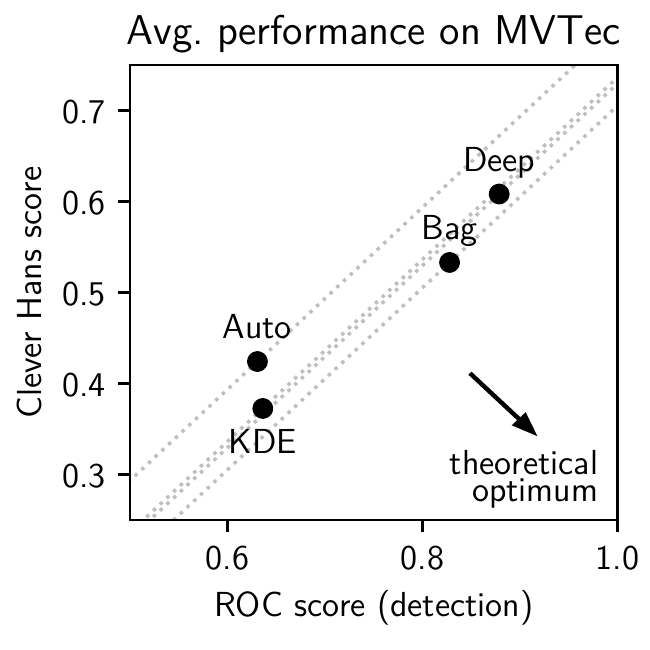}
\vspace{-4mm}
\end{minipage}%
\begin{minipage}{.65\textwidth}
\flushright\parbox{.15\textwidth}{\centering \scriptsize \sffamily  Input}
\parbox{.15\textwidth}{\centering \scriptsize \sffamily ground-truth}
\parbox{.15\textwidth}{\centering \scriptsize \sffamily  KDE}
\parbox{.15\textwidth}{\centering \scriptsize \sffamily  Auto}
\parbox{.15\textwidth}{\centering \scriptsize \sffamily  Deep}
\parbox{.15\textwidth}{\centering \scriptsize \sffamily  \textbf{Bag}}

\vspace{-1mm}

\flushright\fbox{\includegraphics[width=.15\textwidth]{heatmaps/mvtec_ae_bottle_input.png}}
\includegraphics[width=.15\textwidth]{heatmaps/mvtec_ae_bottle_groundtruth.png}
\includegraphics[width=.15\textwidth]{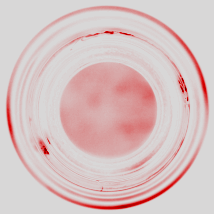}
\includegraphics[width=.15\textwidth]{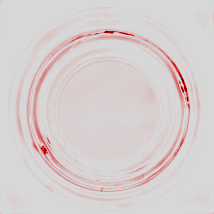}
\includegraphics[width=.15\textwidth]{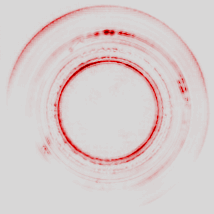}
\includegraphics[width=.15\textwidth]{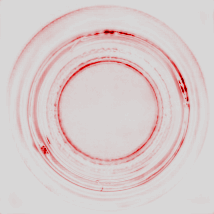}

\vspace{-1mm}

\flushright\fbox{\includegraphics[width=.15\textwidth]{heatmaps/mvtec_kde_wood_input.png}}
\includegraphics[width=.15\textwidth]{heatmaps/mvtec_kde_wood_groundtruth.png}
\includegraphics[width=.15\textwidth]{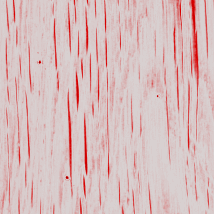} 
\includegraphics[width=.15\textwidth]{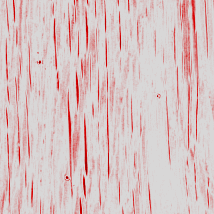} 
\includegraphics[width=.15\textwidth]{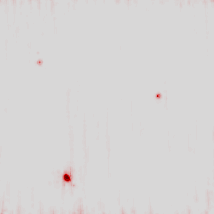}
\includegraphics[width=.15\textwidth]{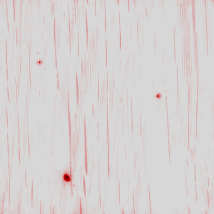}

\vspace{-1mm}

\flushright\fbox{\includegraphics[width=.15\textwidth]{heatmaps/mvtec_nn_zipper_input.png}}
\includegraphics[width=.15\textwidth]{heatmaps/mvtec_nn_zipper_groundtruth.png}
\includegraphics[width=.15\textwidth]{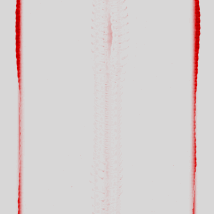} 
\includegraphics[width=.15\textwidth]{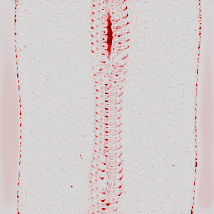} 
\includegraphics[width=.15\textwidth]{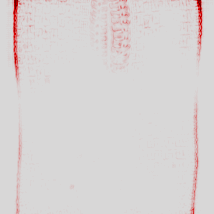}
\includegraphics[width=.15\textwidth]{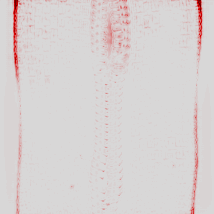}
\end{minipage}
\caption{Left: ROC detection performance and Clever Hans score for each model, averaged over all classes of the MVTec dataset. Right: Examples of explanations for each model.}
\label{figure:bagging}
\end{figure}

We test the bagging approach on the MVTec dataset. Using the bagged outlier score obtained from averaging the standardized scores of individual models, we again compute the ROC detection score for each class. Because the bagged model simply adds an average pooling layer, it still fits in the 3-layer feature\,/\,distance\,/\,pooling architecture presented in Section \ref{section:models}, and we can therefore also compute the explanations and the Clever Hans score. ROC and Clever Hans scores averaged over all classes are plotted for each model in Figure \ref{figure:bagging} (left).

Looking only at the ROC score, the bagged model appears to be outperformed by the deep model. However, a closer inspection reveals that the higher measured detection performance of the deep model comes along with stronger Clever Hans effects. Taking both factors into consideration, the bagged model ranks first among all four models. Figure \ref{figure:bagging} (right) shows explanations produced by each model on the same images as in Fig.\ \ref{figure:heatmaps}. We observe that although relatively far from the ground-truth, explanations for the bagged model comprise a broader range of feature types compared to individual models, hence, pixels used for the decision more frequently overlap with the truly anomalous pixels. This wider support for the detection task should consequently translate into a better generalizing model.

\section{Conclusion}

The `Clever Hans' effect has been often observed in the context of supervised classifiers. In this work, we demonstrated that this effect also occurs in unsupervised learning, specifically, in anomaly detection. Through a newly contributed XAI procedure highlighting relevant features for each anomaly model and a way of quantifying the Clever Hans effect based on matching pixel-wise ground-truth annotations, our work has revealed the widespread occurrence of Clever Hans phenomena in anomaly detection models, additionally exhibiting a wealth of forms such effect can assume in practice. Furthermore, our analysis has revealed that the Clever Hans effect can be mainly attributed to the {\em structure} of the anomaly detection models rather than the data itself.---In fact, the same could be said about the original `Clever Hans' horse when faced with arithmetic calculations: Even if the horse would have seen many mathematical formulas along with their correct answer (including cases where the trainer was not there!), the horse would still be unable to learn the proper problem representation as he is structurally unable to do so. He would therefore invariably continue to predict as a `Clever Hans'.

Interestingly, {\em every} anomaly detection model in our study exhibits Clever Hanses reasoning at least in some cases, and each of them does so in its own particular way. While our work warns against an unreflected use of anomaly detection models in practice, especially for safety-critical tasks, it also sheds a more optimistic note on the problem, specifically, we have demonstrated that a simple bagging approach combining the various Clever Hans models can reduce the Clever Hans effect and lead to sensibly improved results. Future work will aim to go beyond simple bagging to develop new and structurally less rigid models in order to avoid Clever Hans strategies and further improve generalization performance.

%

\section*{Acknowledgments}
This work was funded by the German Federal Ministry of Education and Research (BMBF) as BIFOLD: Berlin Institute for the Foundations of Learning and Data (ref.\ 01IS18025A and ref.\ 01IS18037A) and ALICE III (01IS18049B), as well as by the German Research Foundation (DFG) as Math+: Berlin Mathematics Research Center (EXC 2046/1, project-ID: 390685689). 
This work was partly supported by the Institute for Information \& Communications Technology Planning \& Evaluation (IITP) grant funded by the Korea government (No.\ 2017-0-00451, No.\ 2017-0-01779).

\bibliographystyle{abbrvnat}
\bibliography{bibliography.bib}

\clearpage

\appendix
\setlength{\fboxsep}{0.5em}  

\section{Details on LRP-Based Anomaly Explanation}
\label{appx:explain_details}

In the following, we give some background on the layer-wise relevance propagation (LRP) \cite{bach-plos15} approach to explanation, and more specifically, details and justification for the procedure we use in the paper to explain anomalies. LRP is an explanation technique that leverages the internal structure of neural networks to ease the process of explanation. LRP operates by performing a purposely designed backward pass in the neural network from the output to the input. The backward pass can be derived from the framework of deep Taylor decomposition (DTD) \cite{DBLP:journals/pr/MontavonLBSM17}. The LRP method was later on extended to other models, e.g.\ kernel-based anomaly detection \cite{DBLP:journals/pr/KauffmannMM20}, where one first needs to perform a preliminary `neuralization' step which transforms the model into an equivalent neural network. In the present work, we have generalized the approach to a broader family of anomaly detection models. The neural network equivalents of considered models are shown in Table \ref{table:neuralization}.

\begin{table}[h]
\centering
\begin{tabular}{r|l}\toprule
\textbf{KDE} & Squared Distance $\to$ Negative LogSumExp\\
\textbf{Auto} & Squared Distance\\
\textbf{Deep} & Linear/ReLU $\to$ $\dots$ $\to$ Linear/ReLU $\to$ Linear $\to$ Squared Distance\\
\textbf{Bagged} & [KDE~|~Auto~|~Deep] $\to$ Average Pooling\\\bottomrule
\end{tabular}

\medskip

\caption{Anomaly detection models and their neural network equivalents.}
\label{table:neuralization}
\end{table}

\begin{figure}[h]
\centering
\includegraphics[width=.8\textwidth]{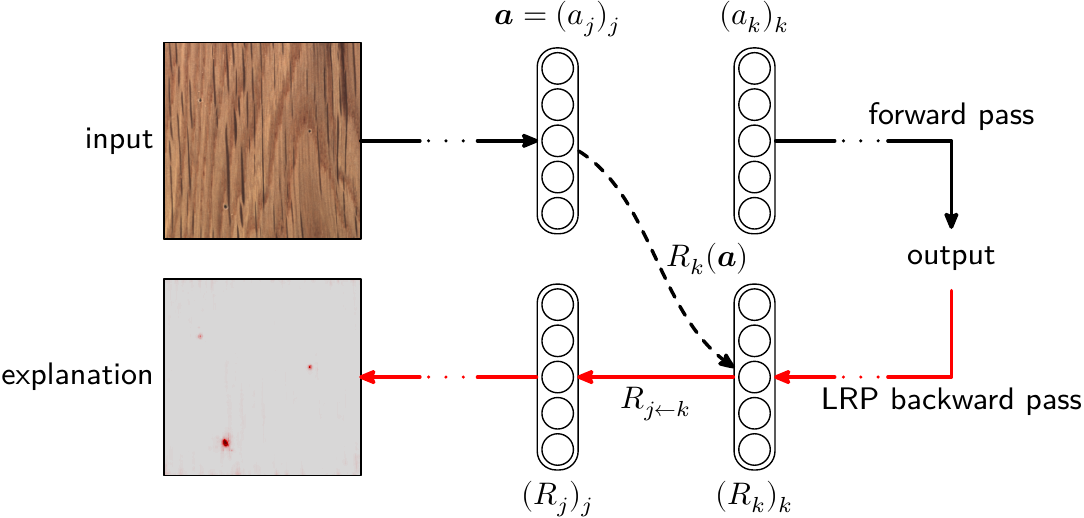}
\caption{Process of prediction of some input data by a neural network, and explanation.}
\label{figure:nn}
\end{figure}

Figure \ref{figure:nn} illustrates the LRP approach. The derivation of the backward pass with DTD is based on a relevance function $R_k(\ba)$ and its Taylor expansion gives the messages $R_{j \gets k}$ to propagate backwards.

Because the function $R_k(\ba)$ can be complex, it is typically substituted by a {\em relevance model} $\widehat{R}_k(\ba) = a_k(\ba) \cdot c_k$ with $c_k$ constant. Thus, in the following, we discuss for each layer type encountered in our anomaly detection models:
\begin{enumerate}
\item For the analyzed relevance model, how to choose an appropriate root point $\rba$ at which to perform the Taylor expansion so that meaningful messages $R_{j \gets k}$ can be extracted.
\item Whether the redistributed messages $R_{j \gets k}$ (or their aggregation $R_j = \sum_k R_{j \gets k}$) are structured in a way that an appropriate relevance model can be built to iterate the propagation procedure one layer below. 
\end{enumerate}

We start the discussion with the pooling layers, and then continue with the distance layer and finally the feature layers of the deep network.

\subsection{Propagation in Average Pooling Layers}

Average pooling layers are defined as:
$$
a_k(\ba) = \frac1N \sum_{j=1}^N a_j
$$
where $N$ is the number of neurons in the pool. We assume that the relevance $R_k$ attributed to the output of the pool is appropriately described by the relevance model:
\begin{align*}
\widehat R_k(\ba) &= a_k(\ba) \cdot c_k\\
 &= \Big(\frac1N \sum_{j=1}^N a_j\Big) \cdot c_k
\end{align*}
with $c_k$ constant. The relevance model can be expanded at the root point $\widetilde\ba = \boldsymbol{0}$, which gives us the first-order terms:
\begin{center}
\fbox{$
\displaystyle
R_{j \gets k} = \frac{c_k}{N} \cdot a_j
$}
\end{center}
defining how much of the relevance in $R_k$ must be redistributed on the neurons of the layer below.

\medskip

As a second step, we would like to verify that the lower-layer relevances are structured in a way that they support a suitable model for propagation in the layer below. Here, we note that
$$
R_{j \gets k} = a_j \cdot {\underbrace{c_k / N}_{c_j}}
$$
where $c_j$ can again be treated as constant.

\subsection{Propagation in Negative Log-Sum-Exp Layers}

We consider a negative log-sum-exp pooling
\begin{align*}
a_k(\ba) = {\smin_j}^\gamma\{a_j\}
\end{align*}
where we have used the notation ${\smin}_j^\gamma\{ a_j \} = -\gamma^{-1}\log \sum_{j=1}^N\exp(-\gamma\, a_j)$, and where $\gamma$ is the stiffness parameter. Like for average pooling, we assume the relevance to redistribute can be modeled as
\begin{align*}
\widehat R_k(\boldsymbol{a}) =  a_k(\ba) \cdot c_k
\end{align*}
with $c_k$ constant. A first-order Taylor expansion at the root point $\rba = \ba - a_k \boldsymbol{1}$ gives the first-order terms \cite{DBLP:journals/pr/KauffmannMM20}:
\begin{center}
\fbox{$\displaystyle
R_{j\leftarrow k} = {\sargmin}_j^\gamma \{a_j\}\, R_k,
$}
\end{center}
where we have used the notation $\sargmin_j^\gamma\{a_j\} = \exp(-\gamma\cdot a_j) / \sum_{j'}\exp(-\gamma\cdot a_{j'})$. This expression can be further developed to let appear $a_j$ as a factor:
\begin{align*}
R_{j \gets k} = a_j \, {\underbrace{{\sargmin}_j^\gamma \{a_j\} \, c_k}_{p_k}} + {\underbrace{{\smin}_{j'}^\gamma \{ a_{j'} - a_j\} \, {\sargmin}_j^\gamma \{a_j\} \, c_k}_{\theta_k}}.
\end{align*}
When $\gamma \to \infty$ (and assuming that elements in the pool have different values), the term $p_k$ becomes locally constant and the $\theta_k$ converges to zero (a proof is given in \cite{DBLP:journals/pr/KauffmannMM20}). This limit result gives support for treating the two terms as constant and zero respectively when propagating to the lower layers, and therefore, a suitable relevance model can be built for the layer below.

\subsection{Propagation in Squared Distance Layers}
Distance layers considered in the main text have the form
\begin{align*}
a_k(\ba) = \|\ba - \boldsymbol{\mu}_k\|^2
\end{align*}
Again, we assume we can build a relevance model of the type $\widehat R_k(\ba) =  a_k(\ba) \cdot c_k$
with $c_k$ constant, and we also treat ${\boldsymbol{\mu}}_k$ to be constant. Here, the function is quadratic, hence, decomposition can only be achieved by performing a second-order Taylor expansion. Choosing the root point $\widetilde{\boldsymbol{a}} = \boldsymbol{\mu}_k$ yields the second-order terms \cite{DBLP:journals/pr/KauffmannMM20}:
\begin{center}
\fbox{$\displaystyle
R_{j\leftarrow k} = c_k \,(a_j - \mu_{jk})^2$}
\end{center}
When each input feature contributes to several distances (e.g.\ the distances to the multiple training points in KDE), relevance scores can be aggregated as: $R_j = \sum_{k}R_{j\leftarrow k}$. When there is only a single distance and this distance is w.r.t.\ the origin (as for the deep model considered in the main text), the equation above reduces to $R_j \leftarrow c_k a_j^2$. In the case of the deep model, the relevance score needs to be further propagated through the multiple layers of features.

\subsection{Transition from Distance to Feature Layers}\label{section:transition}
The last layer of features in the deep model is a linear whitening layer, whose output neurons can be simply written as:
$$
\textstyle a_k(\ba) = \sum_j a_j w_{jk}
$$
Relevance coming from the distance layer above has the form $R_k(\ba) = (a_k(\ba))^2 c_k$ with $c_k$ approximately locally constant. Hence, combining the two equations, we can build the relevance model
\begin{align*}
\widehat R_k(\ba) &= \textstyle \big(\sum_j a_j w_{jk}\big)^2\, c_k
\end{align*}
with $c_k$ constant. However, because of the squaring operation, it is difficult to extract simple messages $R_{j\gets k}$ to redistribute to the layer below. Instead, we observe that the relevance model can be decomposed into an infinite sum of piecewise linear relevance models:
\begin{align*}
\widehat{R}_{k\tau}(\ba) &= \textstyle \max(0,\sum_j a_j w_{jk} \, \mathrm{sign}(a_k) - \tau) \cdot 2 \Delta \tau \cdot c_k
\end{align*}
with small intervals $\Delta\tau$. These relevance models jointly sum to the original relevance model (i.e.\ $\sum_{\tau \in \{0,\Delta\tau,2\Delta\tau,\dots\}} \widehat{R}_{k\tau}(\ba) \approx \widehat{R}_k(\ba)$). For all of these models, there is a root point on the interval $[\boldsymbol{0},\ba]$, and choosing $\rba$ very close to the root point but still on the activated domain gives the directional redistribution:
$$
R_{j \gets k\tau} = \frac{a_j w_{jk}}{\sum_{j'} a_{j'} w_{j'k}} R_{k\tau}
$$
Finally, summing over $\tau$ gives:
\begin{center}
\fbox{$\displaystyle
R_{j \gets k} = \frac{a_j w_{jk}}{\sum_{j'} a_{j'} w_{j'k}} \,R_k
$}
\end{center}
As a last step, we need to verify that the scores $R_{j \gets k}$, or more precisely, the aggregated score $R_j = \sum_k R_{j \gets k}$ support an appropriate relevance model for the layer below. We observe that the relevance can be restructured as:
\begin{align*}
R_{j} &= a_j \cdot \Big( \sum_{k}  w_{jk} \frac{\big(\sum_{j'} a_{j'} w_{j'k}\big)^2}{\sum_{j'} a_{j'} w_{j'k}} c_k \Big)\\
 &= a_j \cdot \Big( \sum_{k} w_{jk} \Big(\sum_{j'} a_{j'} w_{j'k}\Big)\, c_k \Big)\\
&= a_j c_j
\end{align*}
where we have $a_j$ appearing as a linear term, and $c_j$ a term that depends on $a_j$ but only through a nested sum involving many other activations, hence diluting the dependency.

\subsection{Propagation in Feature Layers}
Deep neural networks used in this work are composed of a succession of Linear/ReLU layers of the type
\begin{align*}
a_k(\ba) &= \textstyle \max\big(0,\sum_j a_j w_{jk}\big)
\end{align*}
Again, we assume the model is given as a multiple of the output activation, i.e.\ 
\begin{align*}
\widehat R_k(\ba) &= a_k(\ba) \cdot c_k\\
&= \textstyle \max\big(0, \sum_j a_j w_{jk}\big) \cdot c_k.
\end{align*}
The function is linear with $\ba$ on the activated domain. A root point can be found on the segment
$$
\rba \in \{\ba - t \cdot \ba \odot(\boldsymbol{1} + \gamma \cdot \boldsymbol{1}_{\boldsymbol{w}_{k} \succeq \boldsymbol{0}}) \mid t \in \mathbb{R}\}
$$
where $\gamma$ is a hyper-parameter between $0$ and $\infty$, and $\boldsymbol{1}_{\{\cdot\}}$ is an indicator function applied element-wise. Performing a Taylor expansion at this root point, more exactly, very close to the root point but still in the activated domain, gives the first-order terms \cite{DBLP:series/lncs/MontavonBLSM19}:
\begin{center}
\fbox{$\displaystyle
R_{j\leftarrow k} = \frac{a_j (w_{jk} + \gamma w_{jk}^+)}{\sum_{j'} a_j (w_{j'k} + \gamma w_{j'k}^+)} R_k.$}
\end{center}
The higher the parameter $\gamma$, the more preference is given to the positive contributions to support the explanation. Empirically, this also leads to explanations that are more robust to the high nonlinearity \cite{DBLP:series/lncs/MontavonBLSM19}.

Finally, we need to verify that a suitable relevance model can be built for the layer below. We observe that relevance in that layer can be written as
\begin{align*}
R_j &= a_j \cdot\Big( \sum_{k} (w_{jk} + \gamma w_{jk}^+) \, \frac{\max(0, \sum_j a_j w_{jk})}{\sum_{j'}a_{j'}(w_{j'k} + \gamma w_{j'k}^+)} \, c_k\Big)\\
&= a_j c_j
\end{align*}
where similarly to Section \ref{section:transition}, we can use the nested sums argument to justify the treatment of $c_j$ as constant in the relevance model of the layer below. For further details on how to propagate relevance in the various layers of a deep neural network, see \cite{DBLP:series/lncs/MontavonBLSM19}.

\section{Behavior of KDE for High-Dimensional Data}
\label{appx:high-dim_kde}

High-dimensional input spaces are subject to the effect of concentration of distances, where distances between different randomly sampled vectors become similar \cite{DBLP:conf/icdt/BeyerGRS99,DBLP:journals/sadm/ZimekSK12}. Here, we show that under such effect, the KDE model for outlierness becomes approximately a simple distance-to-the-mean function, i.e.\
$$
o(x) \propto \|x-\mu\|^2 + \text{const.}
$$
To show the above, we decompose the distances as:
$$
\|x-x_j\|^2 = \bar{d} + (\|x-x_j\|^2 - \bar{d})
$$
i.e.\ a mean distance (over all data points $(x_j)_{j=1}^N$), and a deviation from the mean, which according to the concentration of distances effect becomes small. Taking the KDE outlier function, and making use of the approximations $\exp(t) = 1+t$ and $\log(N+t) = \log(N) + t/N$, which are valid when $t$ is close to $0$, we start from the KDE model and arrive at the stated reduction.
\begin{align*}
\gamma^{-1} o(x) &= -\frac1\gamma \log \sum_j \exp (- \gamma \|x-x_j\|^2)\\
 &=  \bar{d} -\frac1\gamma \log \sum_j \exp (- \gamma (\|x-x_j\|^2 - \bar{d}))\\
 &\approx  \bar{d}  -\frac1\gamma \log \sum_j (1 - \gamma (\|x-x_j\|^2 - \bar{d}))\\
  &=  \bar{d} -\frac1\gamma \log (N - \gamma \sum_j (\|x-x_j\|^2 - \bar{d}))\\
    &\approx  \bar{d} -\frac1\gamma \Big[ \log (N) - \frac1N \gamma \sum_j (\|x-x_j\|^2 - \bar{d}))\Big]\\
    &=  -\frac1\gamma \log(N) + \frac1N \sum_j \|x-x_j\|^2\\
    &= -\frac1\gamma \log(N) + \|x-\mu\|^2 - \|\mu\|^2 + \frac1N\sum_j \|x_j\|^2\\
    &= \|x-\mu\|^2 + \text{const.}
\end{align*}

\end{document}